\documentclass[10pt,twocolumn,letterpaper]{article}

\usepackage{titling}
\usepackage{wacv}
\usepackage{times}
\usepackage{epsfig}
\usepackage{graphicx}
\usepackage{amsmath}
\usepackage{amssymb}
\usepackage{booktabs}
\usepackage[accsupp]{axessibility}
\usepackage[hyphens]{url}
\usepackage{pdfpages} 

\def\wacvPaperID{836} % Enter the WACV Paper ID here

\wacvapplicationstrack % Uncomment this line if you are submitting to the Applications Track.

\wacvfinalcopy % *** Uncomment this line for the final submission

\ifwacvfinal
\usepackage[breaklinks=true,bookmarks=false]{hyperref}
\else
\usepackage[pagebackref=true,breaklinks=true,colorlinks,bookmarks=false]{hyperref}
\fi

\begin{document}

%%%%%%%%% TITLE
\title{My Face My Choice: Privacy Enhancing Deepfakes for \\ Social Media Anonymization}

\author{Umur A. \c{C}ift\c{c}i\\
Binghamton University \\
{\tt\small uciftci@binghamton.edu}
\and
Gokturk Yuksek \\
Binghamton University \\
{\tt\small gokturk@binghamton.edu}
\and
\.Ilke Demir \\
Intel Labs \\
{\tt\small ilke.demir@intel.com}
}

\maketitle
  \begin{figure*}[ht]
\begin{center}
    %\centering
    \includegraphics[width=0.85\textwidth]{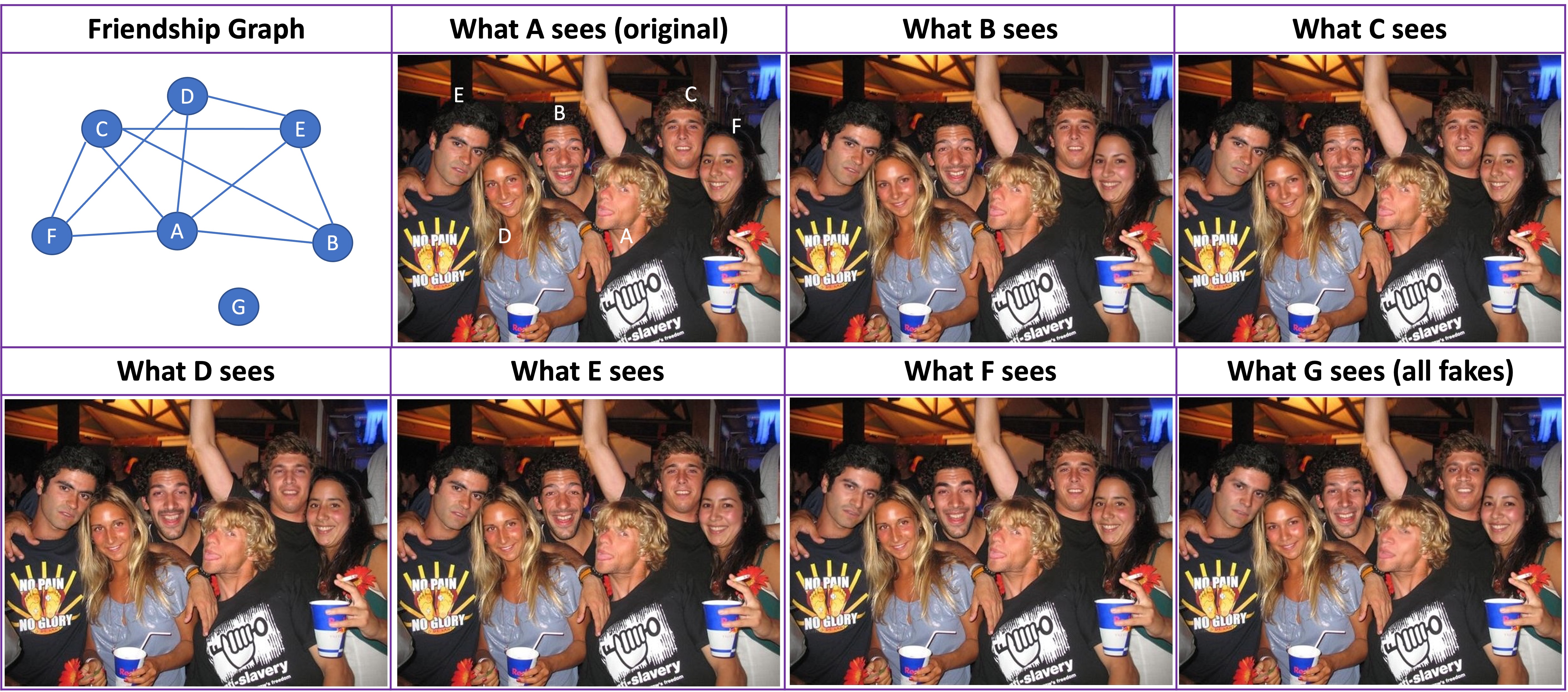}
\end{center}
    \caption{Our anonymization system masks faces with quantitatively dissimilar deepfakes in social photos, according to the friend graph and access rights. Here, A uploads a photo, B-F sees versions where non-friend faces are faked, and G sees everyone as fakes.}
    \label{fig:teaser}
        \end{figure*}
        
\thispagestyle{empty}

%%%%%%%%% ABSTRACT
\begin{abstract}
Recently, productization of face recognition and identification algorithms have become the most controversial topic about ethical AI. As new policies around digital identities are formed~\cite{digitalid}, we introduce three face access models in a hypothetical social network, where the user has the power to only appear in photos they approve. Our approach eclipses current tagging systems and replaces unapproved faces with quantitatively dissimilar deepfakes. In addition, we propose new metrics specific for this task, where the deepfake is generated at random with a guaranteed dissimilarity. We explain access models based on strictness of the data flow, and discuss impact of each model on privacy, usability, and performance. We evaluate our system on Facial Descriptor Dataset~\cite{tanisik2016facial} as the real dataset, and two synthetic datasets with random and equal class distributions. Running seven SOTA face recognizers on our results, MFMC reduces the average accuracy by 61\%. Lastly, we extensively analyze similarity metrics, deepfake generators, and datasets in structural, visual, and generative spaces; supporting the design choices and verifying the quality.
\end{abstract}

%%%%%%%%% BODY TEXT
\section{Introduction}
Face recognition and identification have been one of the most interesting research topics in computer vision~\cite{zhao2003face}. Although the research has contributed to the collective knowledge; the applications have been controversial because of maleficent motivations, deployment of immature products, and bias in data and algorithms. Furthermore, consequences of these faulty products mostly remained unpunished due to the lack of cyber laws around digital identity. Early Google products tagging some skin tones as non-human~\cite{gorilla}, Meta being forced to delete 2 billion face embeddings~\cite{metadeletes}, or ClearView AI aiming to encode every human's face~\cite{clearnews} are just the tip of the iceberg when it comes to facial identity preservation and the impacts of lack thereof.

As a defense mechanism, counter manipulation techniques such as blurring~\cite{vishwamitra2017}, masking~\cite{sun2018cvpr}, and noise addition~\cite{wen2021identitydp} encounter traditional face recognition. Similarly, adversarial generation and confiscation methods~\cite{dong2019efficient,zhong2020towards} are developed for tricking deep learning based face detection and recognition systems. Although effective, these approaches disable face recognition by altering the image content, thus, it is a matter of time that face recognition algorithms are trained and armed against adversarial attacks~\cite{NEURIPS2019_d8700cbd,Liao_2018_CVPR}. Joining these two fronts, our approach can be thought as a based-on-need masking mechanism that does not break the image continuity, and misleads face recognition systems with fake faces, using deepfakes for good.

We would like to demonstrate this privacy enhancing deepfake scenario on a new form of social network. In current social platforms, access rights are defined per image, which friends (or connections) are allowed to see. However, we all appear in hundreds of photos voluntarily or involuntarily, we believe that the access rights should be designed per face, where everyone has freedom over which photos they appear. In current systems, this is poorly handled with tagging/untagging choices, however the photo, and the faces, live on the platform forever even if all faces are untagged. The basic principle behind "My Face My Choice" (MFMC) is that, your face is replaced with a dissimilar enough deepfake in the view of those you do not grant access. We quantitatively analyze and verify that created deepfakes (1) are not similar to the original face by the embedding distance, (2) are not similar to any other face by using synthetic source images, (3) approximate the original age and gender in the image by the selected distance metric, and (4) preserve the original head-pose and expression by the selected generator. Depending on three access levels providing contextual integrity, we propose solutions with different restrictions where the face embedding may only be stored on the client. Our contributions are listed as:
\begin{itemize}
\item A novel privacy enhancing anonymization system with quantitatively dissimilar deepfakes,
\item Quantitative analysis of deepfakes in image, structure, embedding, reconstruction, and generative spaces, and
\item Design and analysis of the system based on face access rights on a social photo sharing network.
\end{itemize}

MFMC can create deepfake versions of photos with more than 20 people, based on complex access rights given by people in the photo. In addition, we sketch these face access priorities in three levels to enable isolation of the face embedding versus detection and elimination of the face embedding. We use a diverse social image database~\cite{tanisik2016facial} for the real dataset, and we evaluate MFMC using two fake datasets (i) 10,000 deepfakes created by StyleGAN~\cite{stylegan2} with random distribution, and (ii) 10,000 deepfakes created by Generated.Photos~\cite{generatedphotos} with equal distribution across skin tones and genders. We analyze our system using four different GANs, with five distance metrics, and against seven face recognition approaches. We believe that MFMC is the first approach to extensively utilize face embeddings to create useful deepfakes as an adversary to face recognition.

%-------------------------------------------------------------------------

\section{Previous Work}
\subsection{Face Recognition and Identification} %Clearview AI AKA Face recognition systems}
As a structured domain with lots of data, faces have been the most interesting playground for detection, recognition, and identification algorithms since Viola and Jones~\cite{viola2001rapid}. Early systems, deploying those algorithms without proper generalization and accuracy guarantees, caused universal consequences as mentioned in the introduction. 

Using a set of known faces from their training process, current state-of-the-art face identification algorithms~\cite{masi2018deep,adjabi2020past,kortli2020face,wang2018additive,liu2017} try to find the identity of the test sample by calculating its deep feature representation. While training, every face identification network is trained with an appropriate loss function such as angular loss~\cite{liu2017} or softmax~\cite{wang2018additive}, that minimizes deep feature distances between faces of the same person and maximizes it between faces with different IDs. During inference, test sample's deep feature representation is computed and used in a one-to-many similarity comparison with known faces to determine its ID. If the distance is smaller than a threshold, face identification network will predict both faces to belong to the same person. 

Parallel to deep learning increasing the power of such face recognition systems~\cite{deepface,we2016m,liu2017,masi2018deep,adjabi2020past,kortli2020face}, novel open-source systems~\cite{openface,deepface,facenet,arcface} and proprietary software~\cite{clearview,microsoftfacerec} emerge for face detection and identification. As these systems become more popular, privacy implications~\cite{oh2016,hukkela2019} and impacts on critical populations~\cite{gendershades,timnit2021,zou2018} of such have also been analyzed closely.

\subsection{Face Generation and Deepfakes}
Generative Adversarial Networks (GANs)~\cite{goodfellow2014generative} has been the stepping stone for creating realistic human face images that are difficult to visually discern from real faces~\cite{nightingale2021synthetic}. Since then, the realism and resolution of the images significantly increased due to changing the generator from getting input latent code only, to the beginning of the network as an input, and then using mapping networks that transform the latent code to transmit into multiple layers of the network~\cite{Karras_2019_CVPR}. In this process, style concept emerged to control these layers with adaptive instance normalization~\cite{dumoulin2016learned,huang2017arbitrary}. Other recent trends in face generation include providing additional noise maps to the generator to increase variability~\cite{shen2020interpreting,stylegan2}. There has also been some GANs that aim to create privacy-preserving faces or face masks~\cite{maximov2020,hukkela2019,gupta2021}.

Recent advances in face generation allowed easier and realistic generation of deepfakes and other facial manipulations. These manipulations and deepfakes can be grouped in four main categories as  (1) novel face synthesis~\cite{Karras_2019_CVPR,stylegan2} including responsible generation~\cite{mixyn}, (2) identity swap~\cite{korshunova2017,ftgan,faceswap,simswap,fsgan}, (3) attribute manipulation~\cite{huang2017arbitrary}, and (4) expression modification~\cite{thies2019deferred}. Our system utilizes the results of (1) as the source for non-existing faces to create deepfakes and the GANs in (2) for creating deepfakes with the same expressions, poses, and attributes.

\subsection{Privacy Preserving Faces}% AI -- de-ID, anonymization}
Overall, face biometrics is one of the most personally identifiable information that is released without proper access rights~\cite{blaz2021}. As an adversary to face recognition, face obfuscation methods such as blocking and blurring~\cite{vishwamitra2017}, noise addition~\cite{wen2021identitydp}, and inpainting~\cite{sun2018cvpr} have been proposed. However, such approaches break the continuity of the image and reveal that the image is obviously altered. In contrast, de-identification and anonymization approaches keep the image intact and modify the face to trick the recognition systems. Such methods are proposed on images~\cite{newton2005} and videos~\cite{gafni2019}; using model-based~\cite{gross2006}, GAN-based~\cite{wu2019}, or hybrid~\cite{sun2018eccv} approaches; for face detection~\cite{shai2006}, action recognition~\cite{ren2018}, and annotation~\cite{shirai2019}. Our approach is also a privacy-preserving system, however it works in a multi-person setting based on access rights and social graphs, bridging the gap between privacy-preserving algorithms and real-world platforms.

\subsection{Adversarial Attacks}
Another perspective in breaking face identification algorithms, especially for deep learning based methods, is adversarial attacks. Similar to the approaches in the previous section, adversarial attacks also change the image content, mostly in the reconstruction and generative spaces, instead of directly swapping the face. These attacks may be image perturbations~\cite{oh2017}, poisoning attacks on training data~\cite{zhu2019}, or cloaking images right after capture~\cite{shan2020}. Algorithmic manipulations and basic adversarial attacks are shorter term solutions for the ever-improving GANs, so we choose the path of ``creating as many fake faces as possible'' to explode the embedding space of face recognition approaches.

\section{Privacy Enhancing Deepfakes}
The main motivation of MFMC is to keep the face images and face embeddings as local as possible, using the social graph and the access rights set by the users. In a nutshell, when an image is uploaded from the client, the \textit{friends} (i.e., users connected to the uploader) are optionally tagged. Remaining faces in the photo are replaced with deepfakes and the image is sent to the server with this metadata. The tagged friends see others based on their friendship and outsiders see everyone as fakes. In this section, we describe the deepfake target selection metrics, deepfake creation process, and the design of face access rights.

\subsection{Target Face Query}
\label{sec:target}
Anonymization of a face through deepfakes requires replacing the source face with another face that fits the same facial frame in a realistic way. The new face needs to be different enough for the anonymization to be successful. In order to define this difference, we need a comparison metric which yields similar enough faces for contextual and visual continuity, and dissimilar enough to confuse face recognition systems. As the image space is prone to misalignment, illumination, and noise, we proceed with the face embedding space, where we also preserve main face attributes such as age and gender. In our framework, we use ArcFace~\cite{arcface} to extract face embeddings with 512 features. ArcFace, using a novel loss function, optimizes feature embeddings to enforce higher similarity for intra-class samples and diversity for the inter-class samples. Moreover, we utilize InsightFace~\cite{insightface} to perform gender and age based classifications to store the similarity in the latent space. Note that this classification is neither exposed, nor used explicitly in our system, it is only stored as a direction in the latent space to define the similarity bounds. Experiments in other metric spaces are documented in Sec.~\ref{sec:metric}.

% In our experiments, people having similar age prediction share similar face complexion and people having similar gender prediction share similar facial hair and makeup conditions. Having match people with similar facial complexion and facial hair generates more realistic looking faces. We also noticed that facial hair on some people move beyond the facial boundaries for the deepfake and no having facial hair on destination face followed by facial hair on outside of the face generate visible artifacts. Having that preliminary filter based on the embedding values of the given gender and age gives us direction and control over these kind of cases. 

Having computed face embeddings, we need a metric to compare two embeddings in order to query the ``best'' synthetic face as a replacement for the original face. We experiment with (1) minimizing the embedding distance for finding the closest face, (2) maximizing the embedding distance for finding the furthest face, (3) minimizing the embedding distance along the age and gender directions in the latent space as for finding the closest face with similar general attributes, (4) maximizing the embedding distance along the age and gender directions in the latent space for finding the furthest face with similar general attributes, (5) randomizing (2) within a plausible threshold to diversify the created samples, and (6) randomizing (4) within a plausible threshold to diversify the created samples. These choices are further evaluated in Sec.~\ref{sec:res}. Based on the motivation of ``creating as many fakes as possible'' the randomization is needed and the dissimilarity threshold ensures that the deepfake is still not recognizable as the source face. Overall, the set of target faces are created as,
\begin{align}
 T(S) &= \{I_i \in I\}  \ \ where \\ \nonumber
 ||E(I_i)&-E(S)|| >  \\ \nonumber 
\max(||E(I_j)-E(S)||) & - \sigma(||E(I_j)-E(S)||) \\
     \forall j &\in I \nonumber
%\frac{1}{N}\sum_{j<N}{||E(I_j)-E(S)||}  
\end{align}
where $I={I_0,I_1,\dots,I_N}$ represents all images in the synthetic dataset, $S$ is the real source image, and $T(S)$ is the target image set. $||.||$ denotes $\ell_2$ distance, $E(.)$ is the face embedding, and $\sigma$ is the standard deviation.

\subsection{Deepfake Generation}
In our framework, anonymization of a face requires a deepfake technique that fulfills transferring the identity of a synthetic target face into to the original source face while preserving extrinsic attributes such as expression, head pose, gaze direction, and lighting. In addition, the deepfake generator should be able to apply this process to multiple faces independent of occlusions and interactions. 

The traditional deepfake creation with a common encoder and identity-specific decoders is not generalizable for this task as training an identity specific decoder requires multiple images with various poses. Multiple face requirement multiplies the need for compute resources and needs storing face decoders. As storing faces and identities conflicts with our main motivation, we cannot use traditional deepfakes. Overall, the deepfake generator (i) should be generalizable, working with any face (both as a source and a target) without any prior training, (ii) should change the identity of the source with the target significantly, (iii) should preserve facial and environmental attributes such as facial expression, head pose, gaze direction, lighting consistency, and (iv) should not create visible artifacts.

To satisfy these requirements, we integrated multiple deepfake generators into our framework. For 3D face model based approaches, Nirkin et al.~\cite{faceswap} morph a 3D face model created from a source image into target face's expression values and use Poisson Blending to merge the two models. FTGAN~\cite{ftgan} applies Few-Shot Unsupervised Image-to-Image Translation~\cite{liu2019few} and combines it with SPADE~\cite{park2019SPADE} module in order to inject semantic priors for face-swapping. FSGAN~\cite{fsgan} uses a two-stage network architecture. While the first stage network performs the expression transfer with the reenactment process, second stage blends the result of first stage network into target image using face inpainting. Lastly, we integrated SimSwap~\cite{simswap}, which is a modification of the traditional deepfake method with an ID injection module for generalization to arbitrary faces. It also uses weak feature matching loss to realistically reflect the facial attributes while still preserving the source identity information. We evaluated these methods (Sec.~\ref{sec:gans}) and concluded that SimSwap provides the best results on different datasets, measured by five different losses.

\subsection{Face Access Models}

The ability to generate deepfakes must be accompanied by a capable
access model to strike a proper balance between the desire to
participate in social media and to preserve individual
privacy. Furthermore, it needs to accommodate varying degrees of
privacy needs, instead of imposing a one-size-fits-all solution. We assume that social platform operators are willing
participants in MFMC with incentives to improve user privacy and abide
by their privacy guidelines. % Additionally, we trust them to
% generate deepfakes correctly, and remove original images along with
% relevant deepfake metadata when removal is possible. At the expense of
% convenience, which may limit user adoption, the deepfake generation
% can be performed by a trusted third party service instead.
% However, this work currently does not
% incorporate the third party trust model. Further research is required
% to study the feasibility of proposed deepfake face generation on a
% smartphone device as an alternative.
MFMC supports the following access models:

\subsubsection{Disclosure by Proxy} In this model, all the faces in a
picture except the ones chosen by the individual that uploads it to
the social media platform are deepfakes. This is disclosure by proxy
because anybody who is in the picture and wishes to reveal their real
face must be allowed by the original submitter. Once the picture with
real and deepfake faces are finalized, the client disposes of the
original picture. %Alternatively, the generation process can be done on the client side without even pushing the original faces to the server side.

The privacy of individuals depend on the submitter 
respecting others' consent, or lack thereof, to have their real faces
revealed. Since none of the real faces remain
on the platform, this model extends the privacy protection of MFMC
against information leakage where the social media
platform is compromised. In addition, no extra persistent storage is necessary on the client for face embeddings, or on the server as only the deepfake version of a photo is kept.

\begin{figure}[b]
\centering
\includegraphics[width=1\linewidth]{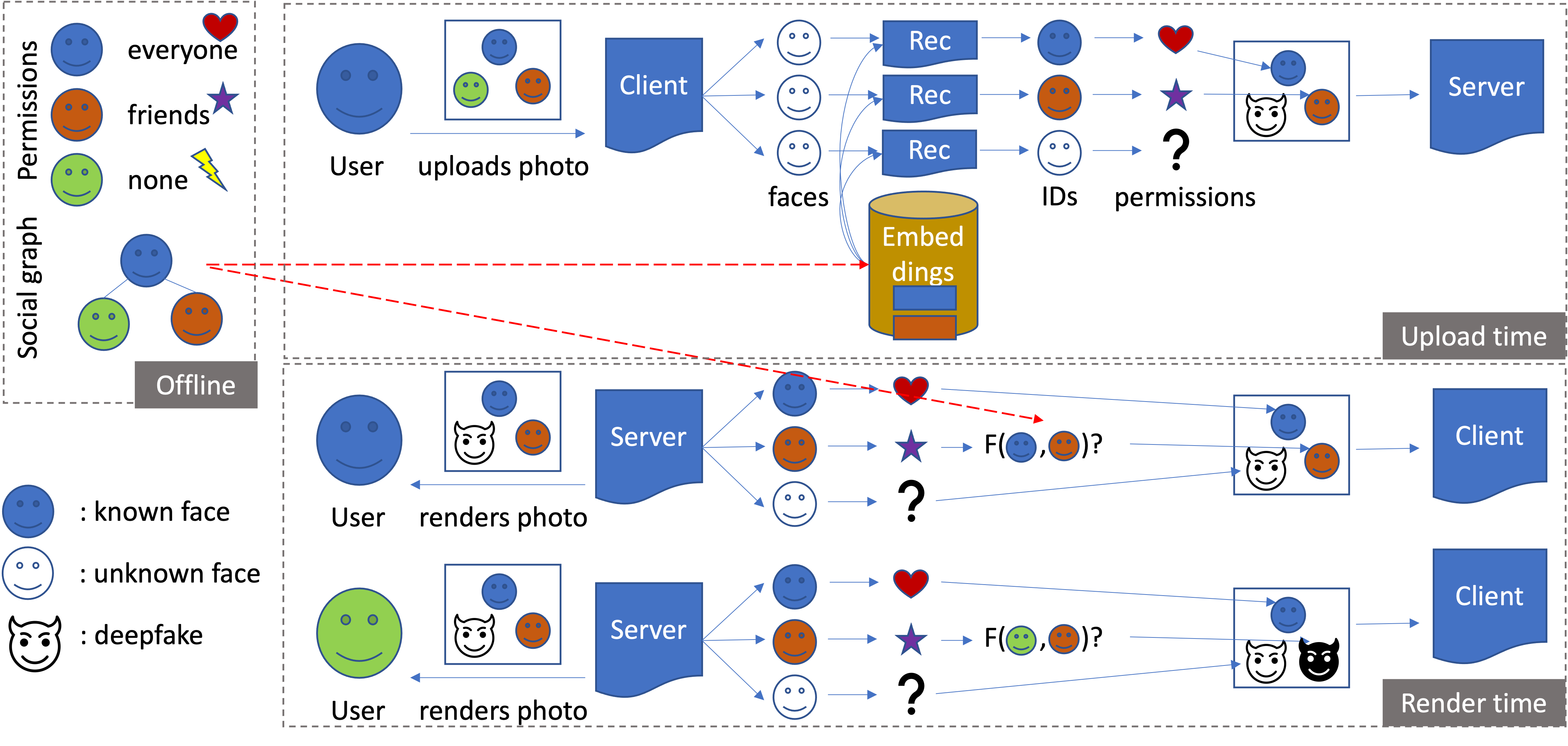} 
\caption{\textbf{Workflow for a three user network.} Green does not allow being seen by anybody, so its face embeddings are never shared, and it is always deepfaked on client. Blue grants to be seen by everyone, so it is always real. Orange only wants to be recognized by friends, so its face embedding is only shared with friends, and is deepfaked by the server at render time, per viewer. }
\label{fig:workflow}
\end{figure}

\begin{figure*}[h!]
\centering
\includegraphics[width=0.9\linewidth]{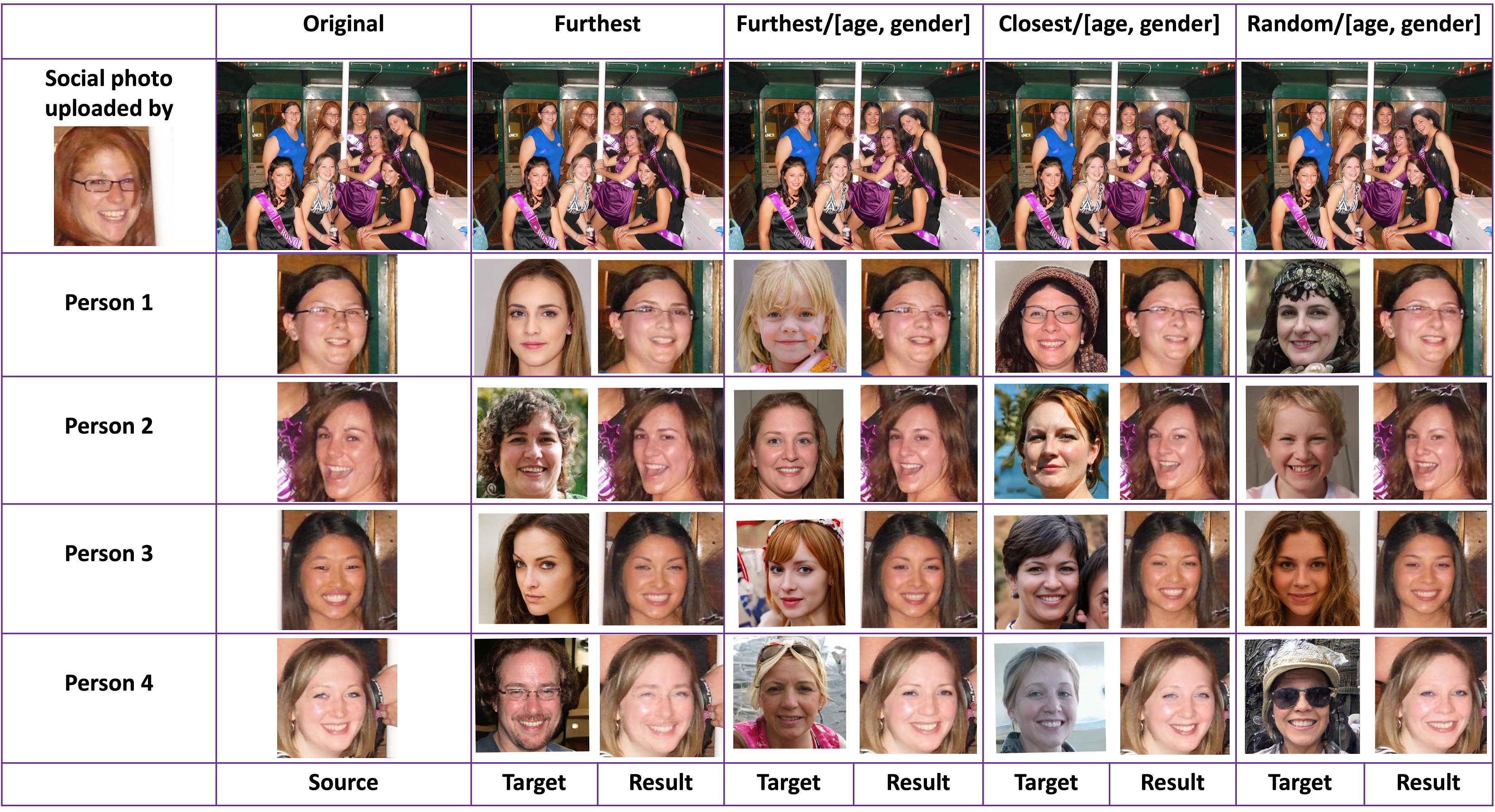} 
\caption{\textbf{MFMC Results per Similarity Query.} Source images and corresponding deepfakes created from closest/furthest/random target queries, and restricted by age and gender groups, are demonstrated.}
\label{fig:main}
\end{figure*}

\subsubsection{Disclosure by Explicit Authorization} This access model
incorporates the tagging feature available in most social media
platforms, in which tagging establishes
a link between the face and a user. Typically, the platform then requests the consent of the
tagged person to be associated with the face. We build on
this feature such that when the users consents, deepfake is not created. Likewise, when a tag is removed,
the platform replaces the real face with a deepfake. All non-tagged
faces are deepfakes by default.

Since the platform strictly requires individual consent for a face
to be tagged, accidental disclosure by a proxy is not possible unlike
in the previous access model. Individuals that do not have accounts
on the social media platform, but have faces captured
in a picture with or without consent, remain anonymous as their faces
are always protected by a deepfake. Given the fluid
nature of tag/untag operations, the platform must keep a permanent
copy of all the real faces both to generate deepfakes for and to
restore as necessary.

\begin{figure}[h!]
\centering
\includegraphics[width=1\linewidth]{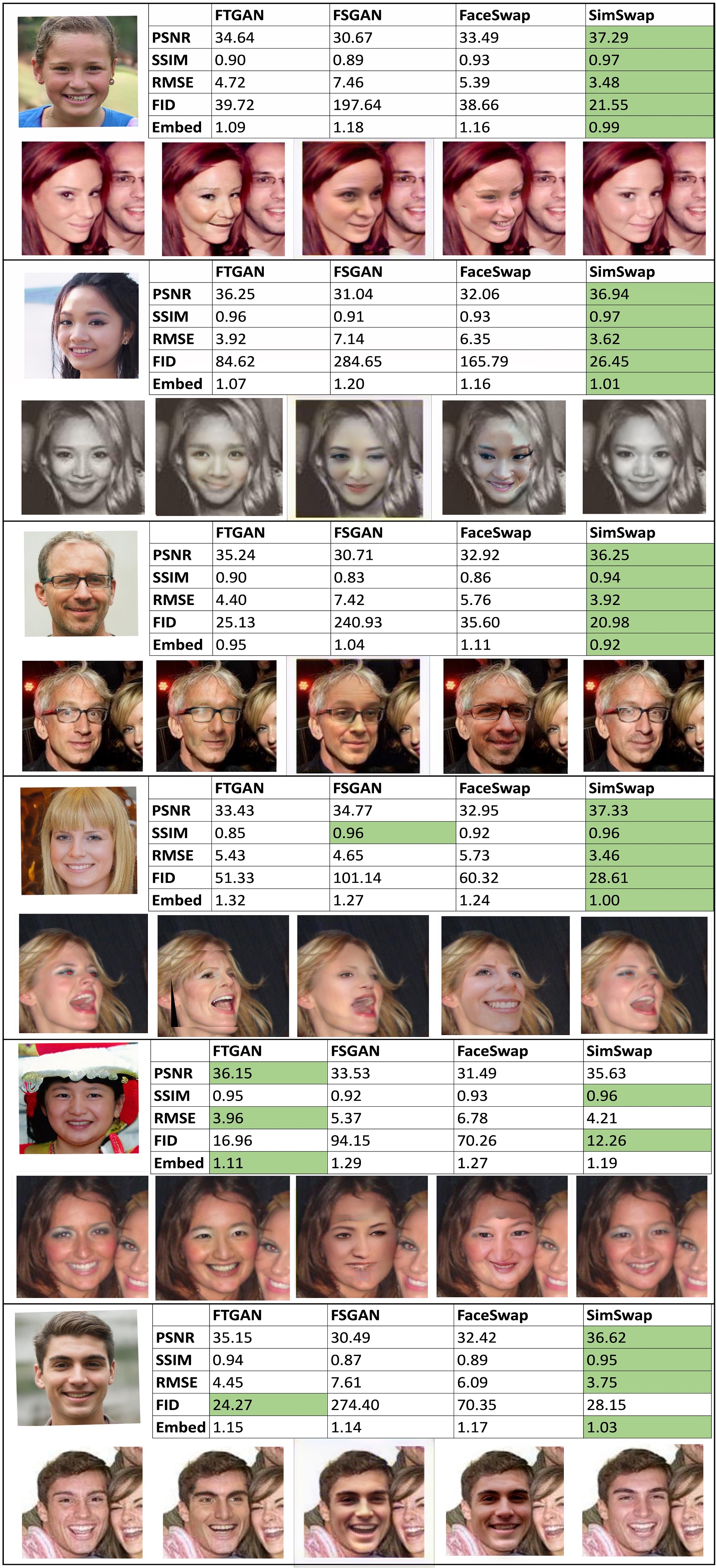} 
\caption{\textbf{GAN Comparison.} Six sample target-source pairs identified by MFMC (left) and corresponding results by 4 GANs are compared in PSNR, SSIM, RMSE, FID, and embedding spaces.}
\label{fig:gans}
\end{figure}

Every tag/untag operation triggers a regeneration of the picture, thus, the performance
requirements are higher compared to the previous model. The storage demands
are also increased due to the need for persistently storing the original
picture as well as the deepfake versions.

\subsubsection{Access Rule Based Disclosure}
Similarly to the previous model, all unknown faces are deepfakes by
default during upload. Face embeddings of permitted friends/followers are kept on device, and if a face embedding is not on the device at upload time, it is unknown. 
% and revealing a face requires its embedding to be on device at upload time, meaning predefined consent of individual associated with the face.
Unlike the previous model, at render time, the
platform consults a set of access rules supplied by face owners, to decide whether their face should be revealed (Fig.~\ref{fig:workflow}).

In their simplest form, access rules would allow a
user to specify how their face should be revealed to friends or
followers versus the general public. More sophisticated access rules
could enable finer-grain control, allow different deepfakes for different observers, and possibly more.

The dynamic nature of different viewers possibly observing different
versions of the same picture comes at the cost of increased storage and
computation demands. More complex access rules
demand more storage and computation.

\section{Results}
\label{sec:res}
%We have used a pre-trained version of arcface~\cite{insightface}
MFMC is implemented using InsightFace~\cite{insightface} library for face detection (which employs a pre-trained version of arcface~\cite{arcface}) and SimSwap~\cite{simswap} library for deepfake generation. We empirically selected InsightFace as a fast and accurate detector over OpenFace~\cite{openface} and FaceNet~\cite{facenet}, however the face detector can be swapped if the speed vs. accuracy trade off changes. Inference and training modules run on an NVIDIA GTX 1060 GPU. 
%For a sample image with 10 people, it takes under a minute to create the version for the first access model (i.e., all deepfakes). 
For the real dataset covering many social interactions in crowded environments, we utilize ``party'' subset of Facial Descriptors Dataset~\cite{tanisik2016facial}. The dataset includes many configurations of 1282 people in 193 images, with various resolutions, poses, and illumination. In order to create the face manifold for MFMC to choose source images from, we use two datasets with unknown and uniform distributions. The first one contains 10,000 fakes created with StyleGAN~\cite{stylegan2}, and the second one contains 10,000 fakes with equal skin tone, gender, and age distributions created by Generated.photos~\cite{generatedphotos}. The similarity metric is face embedding distance unless otherwise is noted.

%Ilkeeee orda misin ? SSIM ve RMSE de closes t mi farthest mi seceyim acc hesaplarken ? ssim buyukse closest, rmse kucukse closest, buna gore furthestlari sec tabi ki... face rec batsin istiyoruz neden closest secelim??!

% Equal diversity	30,000	25\% of each ethnicity, 50/50 genders
% https://generated.photos/datasets 
% 1000 stylegan images
% https://github.com/coallaoh/PIPA\_dataset
% %https://vision.cs.hacettepe.edu.tr/interaction_images/

Fig.~\ref{fig:main} contains MFMC results for an uploaded photo. For each face, we show the chosen target image from StyleGAN dataset and the created deepfake, according to four metrics as explained in Sec.~\ref{sec:target}. As seen in the furthest target image of Person 4, opposite genders may be chosen if there is no age and gender normalization. Similarly, as in the closest target image of Person 1, the deepfake may be too similar. Normalized furthest images with a randomness interval creates quantitatively dissimilar deepfakes as demonstrated in the last column. Note that in this access level, all faces except the owner are replaced, however only four of them are analyzed in detail here. Additional MFMC samples per target query type can be explored in Fig.~\ref{fig:additional}.

\subsection{Experiments} 
The quality of MFMC photos depends on the photorealism of the underlying GAN and the target image datasets. In addition, the similarity metric for target query affects the diversity and credibility of the results. % In this section, we experiment with other GANs and evaluate the impact of the target dataset.

\subsubsection{GAN Comparison}
\label{sec:gans}
In Fig.~\ref{fig:gans}, we compare four face generators (FTGAN~\cite{ftgan}, FSGAN~\cite{fsgan}, FaceSwap~\cite{faceswap}, and SimSwap~\cite{simswap}) with five quality metrics (PSNR, SSIM, and RMSE in image space; and FID and face distance in embedding space), on six MFMC results. On the left of each row, we demonstrate the target from StyleGAN dataset (top) and the source (bottom) from the party dataset, followed by the result created by each GAN and their scores. We observe that SimSwap results score as the most preferable (colored in green). Note that image and face based metrics should be evaluated together, otherwise obvious rotation/cropping artifacts as in the fourth sample may be missed. We emphasize that SimSwap creates deepfakes faithful to source resolution (row 2), source headpose and expression (row 4), target race (row 5), and target accessories (row 3). Extending this comparison to the whole dataset, we conclude that SimSwap generates the highest scoring fakes in all metrics.
\begin{figure}[h]
\centering
\includegraphics[width=1\linewidth]{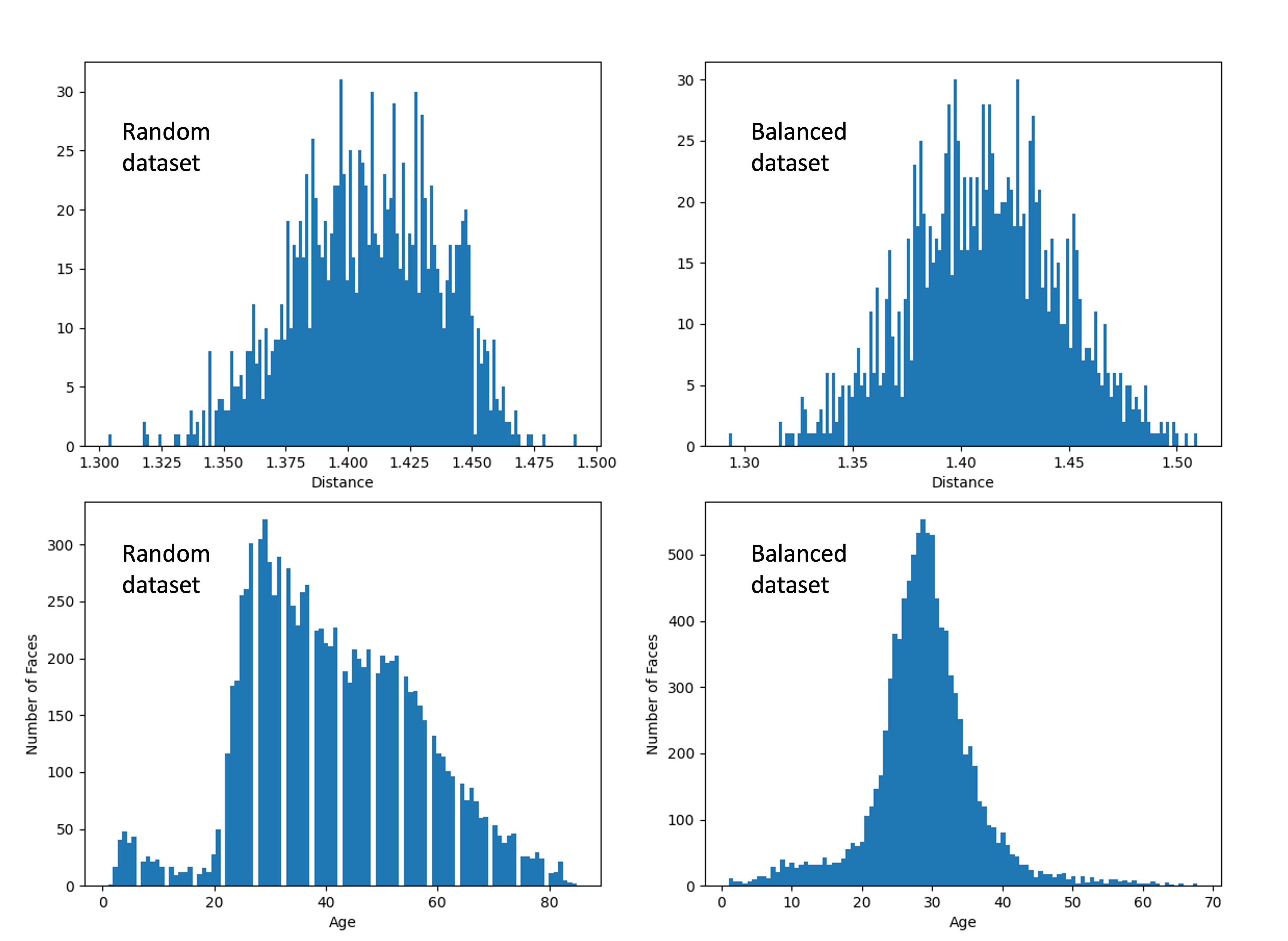} 
\caption{\textbf{Random and Balanced Datasets.} Embedding distance distribution and age distribution of the two datasets are compared.}
\label{fig:GeneratedPhotosEmbedDist}
\end{figure}
\subsubsection{Dataset Dependency}\label{sec:sec412}
As the source repository, we utilize two different synthetic datasets. StyleGAN dataset does not have a predefined distribution of the faces, it contains randomly generated faces in terms of age, gender, and skin tones. On the other hand, Generated.photos dataset has equal distribution in terms of the ranges for the aforementioned categories. This difference is reflected in the distributions depicted in Fig.~\ref{fig:GeneratedPhotosEmbedDist}, and having a uniform range of attributes help MFMC create diverse samples. In contrast, distributions are more similar in terms of embedding distances, which reflects that MFMC is not affected by different datasets for target queries.

%\textbullet All Experiments performed on System setup \SpacingU
%\subsection{Best Deepfake Generator} compare simswap vs. other deepfake generators in terms of ssim rmse fid

%\subsection{Face Embedding size?quality?diversity?}
%\SpacingU\SpacingU\SpacingU\SpacingU\SpacingU
%subsection{Target Face Generator} compare stylegan vs. other target face generators in terms of ssim rmse fid
%\SpacingU\SpacingU\SpacingU\SpacingU\SpacingU
%\subsection{Face Detector/ Age Gender Detector} compare current face detector vs. others in terms of accuracy and attribute overlap
%\SpacingU\SpacingU\SpacingU\SpacingU\SpacingU

\subsubsection{Similarity Metric Selection}
\label{sec:metric}
\begin{figure}[h!]
\centering
\includegraphics[width=1\linewidth]{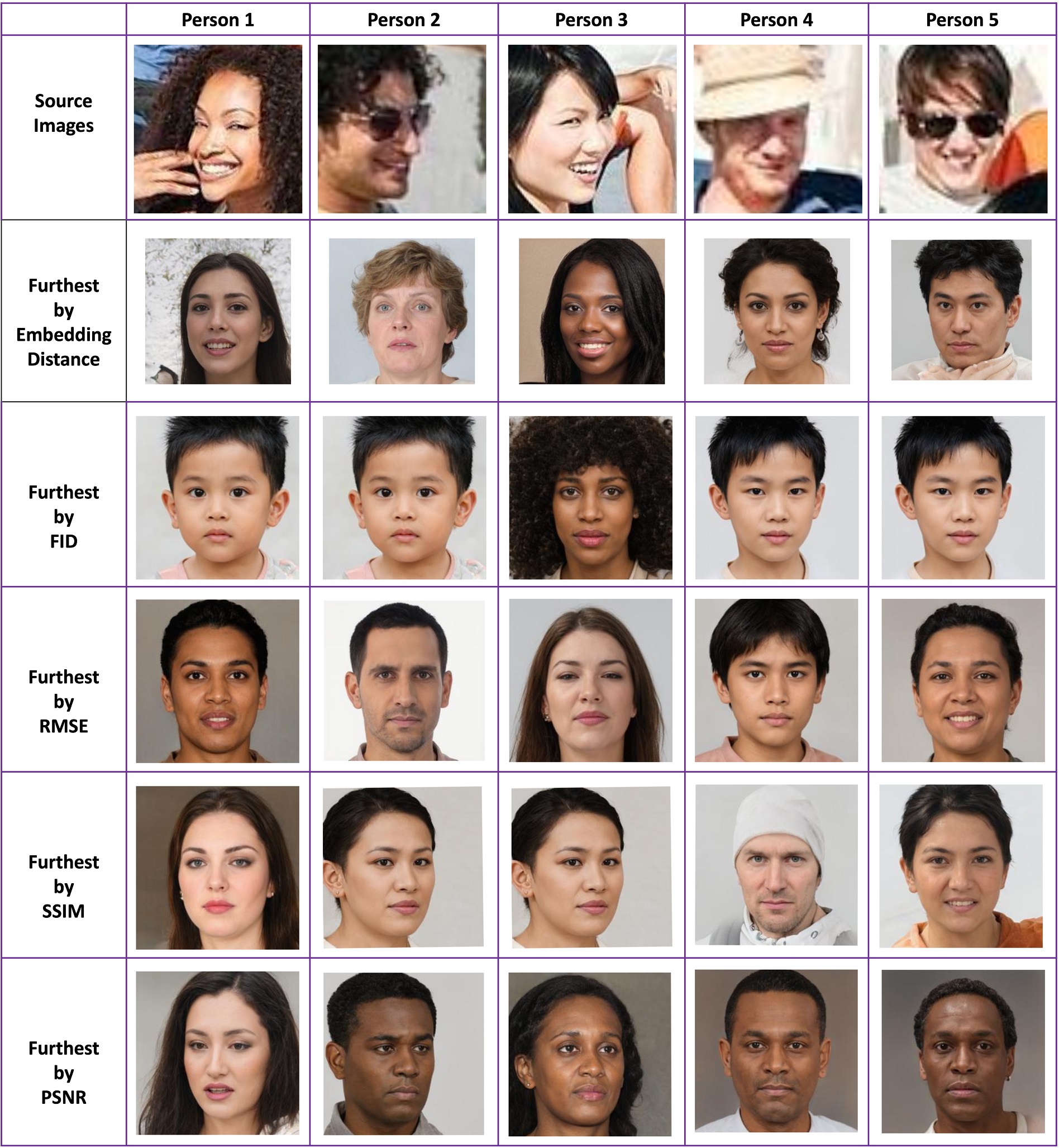} 
\caption{\textbf{Similarity Metric Comparison.} Source (first row) and queried target images per five different metrics are demonstrated.}.
\label{fig:sim}
\end{figure}

We select a sample photo with five people to show the effects of similarity metrics for target image query. Fig.~\ref{fig:sim} gathers furthest target images per source, where the dissimilarity is defined by face embedding distance, FID, RMSE, SSIM, and PSNR in each row. To coherently compare the metrics, (1) we discard the randomness interval by using the most dissimilar and (2) we use the aforementioned equal distribution dataset. Visually and quantitatively the most dissimilar is demonstrated to be the face embedding metric in the first row, whereas FID creates a non-uniform manifold with extrema, RMSE and SSIM weighs headpose and alignment, and PSNR weighs color. 

\subsection{Evaluation} %umur
We evaluate the success of MFMC by the reduction of overall face recognition accuracy and by visual comparison to existing privacy preserving face altercations. %by its expected performance on client and server sides based on the face access models, and by a small user study. 

\begin{table}[ht]
\begin{tabular}{c|cc|cc}\centering
%\multirow{Face} & \multicolumn{2}{c|}{Source - Target} & \multicolumn{2}{c}{Source - Result} \\  \cline{2-5} Detector 
Face & \multicolumn{2}{c|}{Source vs. Target} & \multicolumn{2}{c}{Source vs. Result} \\  \cline{2-5} Detector 
                               & \multicolumn{1}{c|}{Furthest}     & Random    & \multicolumn{1}{c|}{Furthest}    & Random    \\ \hline
FaceNet512                     & \multicolumn{1}{c|}{0.001}        & 0.0      & \multicolumn{1}{c|}{0.14}        & 0.16      \\
OpenFace                       & \multicolumn{1}{c|}{0.001}        & 0.002     & \multicolumn{1}{c|}{0.2}        & 0.23      \\
FaceNet                        & \multicolumn{1}{c|}{0.03}        & 0.02      & \multicolumn{1}{c|}{0.32}        & 0.34      \\
%VGG-Face                       & \multicolumn{1}{c|}{0.04}         & 0.04      & \multicolumn{1}{c|}{0.51}        & 0.61      \\
DLib                           & \multicolumn{1}{c|}{0.02}         & 0.05      & \multicolumn{1}{c|}{0.35}        & 0.45      \\
ArcFace                        & \multicolumn{1}{c|}{0.06}         & 0.04      & \multicolumn{1}{c|}{0.36}        & 0.45       \\ 
DeepID                         & \multicolumn{1}{c|}{0.006}         & 0.01      & \multicolumn{1}{c|}{0.54}        & 0.55      \\
DeepFace                       & \multicolumn{1}{c|}{0.04}         & 0.06      & \multicolumn{1}{c|}{0.57}        & 0.55      \\\hline
Average                        & \multicolumn{1}{c|}{0.02}         & 0.02      & \multicolumn{1}{c|}{0.35}        & 0.39     
\end{tabular}
\caption{\textbf{Face Recognition Accuracies after MFMC.} Seven SOTA approaches are compared on MFMC results based on face identification accuracy, which is reduced by 61\% on the average.}\label{tab:facerec}
\end{table}

\subsubsection{Breaking Face Recognition}
Following the main motivation of MFMC, we want to prevent mainstream face recognition systems from identifying faces. To support this motivation, we test seven state of the art face recognition systems (DeepID~\cite{deepid}, OpenFace~\cite{openface}, DeepFace~\cite{deepface}, FaceNet~\cite{facenet}, FaceNet512~\cite{facenet}, DLib~\cite{dlib}, and ArcFace~\cite{arcface}) on 1282 faces in 193 images created by MFMC. We use per-detector thresholds on the cosine distance between the face descriptors of source vs. target (query synthetic face) and source vs. result (created deepfake) faces from each face detector to decide if they belong to the same person. Tab.~\ref{tab:facerec} documents face recognition accuracies which MFMC reduces by 61\% on the average (last col.), and by 65\% if we lift the randomness concern (fourth col.). The face identification accuracies between source and target faces are reported for validation (second and third cols.), these are expected to be very low as we want justifiably dissimilar deepfakes. For the most popular choice OpenFace, MFMC reduces its accuracy by 80\%. We repeat this experiment using SSIM and RMSE metrics in Supp. B.  %We also note that the accuracy of ArcFace is expected to be higher than others, since it is also used in MFMC.

\begin{figure*}[h!]
\centering
\includegraphics[width=0.95\linewidth]{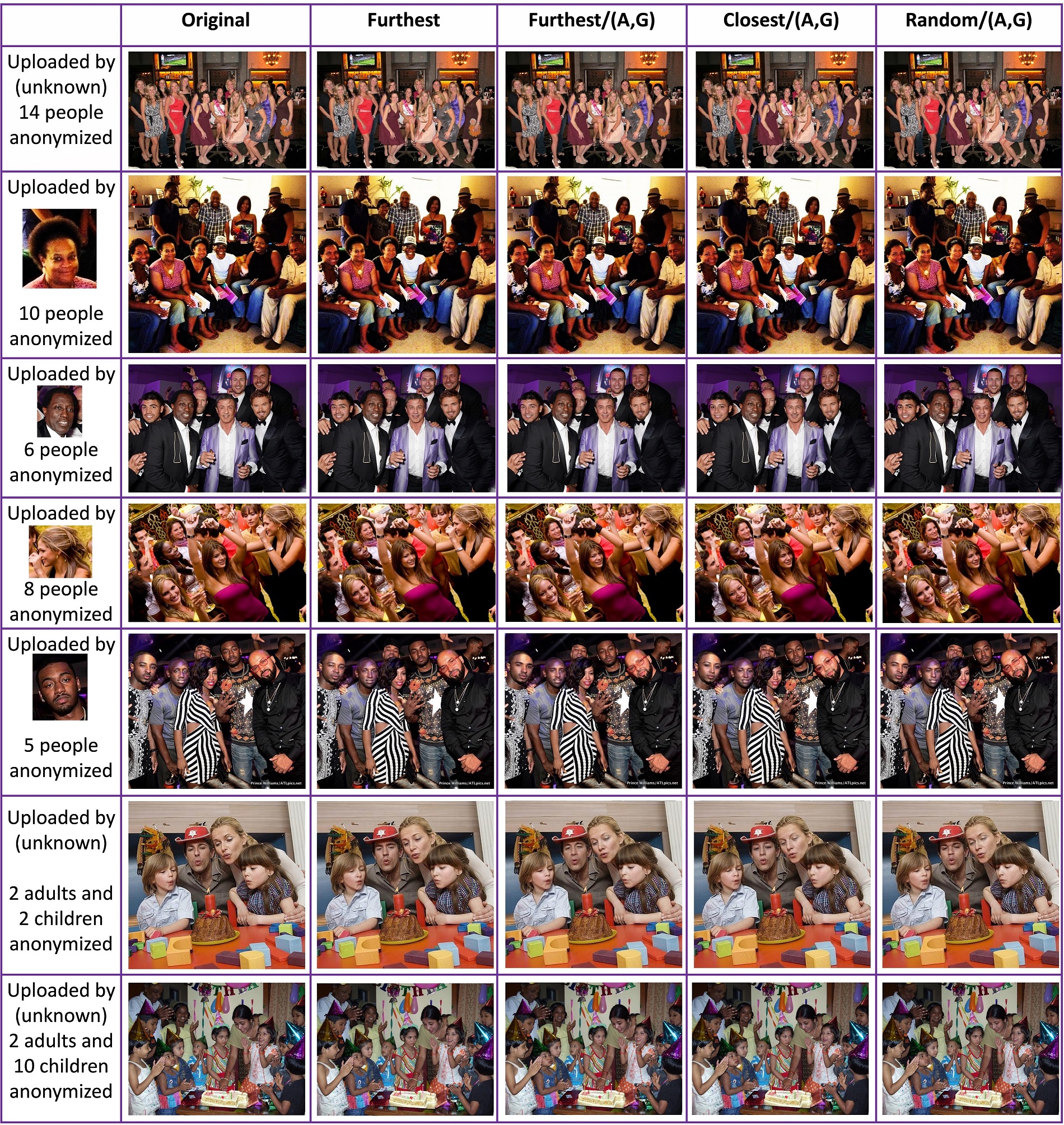} % Reduce the figure size so that it is slightly narrower than the column. Don't use precise values for figure width.This setup will avoid overfull boxes.
\caption{\textbf{Additional MFMC Results.} Original images and privacy enhanced versions based on different target queries are demonstrated. The last column with a randomness threshold from the furthest distance within the same age and gender shows the ultimate results.}
\label{fig:additional}
\end{figure*}

\subsubsection{Comparison}
To the best of our knowledge, MFMC is the first full system design for using deepfakes to solve face ownership problem in social media platforms. Other approaches using masking~\cite{8014915}, filtering~\cite{6918642}, image transformations~\cite{8014911}, inpainting~\cite{hukkela2019} and GANs~\cite{maximov2020} neither design how such algorithms can be systemized for production, nor evaluate against face recognition systems. Other approaches~\cite{proencca2021uu,hukkela2019,chen2021privacy} do not provide control over face parameters, or extreme anonymization obfuscates the face too much that renders the photo meaningless~\cite{yuan2017image,kobayashi2014surveillance,korshunov2013using}. To support this claim, we compare MFMC to CIAGAN~\cite{maximov2020} and Deep Privacy~\cite{hukkela2019} using their sample images in Supp. A.

\section{Conclusion}
We present a privacy-enhancing anonymization system for everyone to have control over their faces in social photo sharing networks, saying \textit{my face, my choice!}. In addition to the new metric for target query for deepfake creation, we extensively analyze different deepfake generators and similarity metrics for this task. MFMC also demonstrates a responsible use for deepfakes by design, especially for protecting faces of minors and vulnerable populations, in contrast to the dystopian scenarios~\cite{ucla}.  We also present different face access models for efficiency and embedding storage. We validate that MFMC is able confuse several current face identifications systems. We believe that current social media platforms would free the users if similar approaches to MFMC are implemented for face privacy and contextual integrity.

As a prototype system, there is always room for improvement. As mentioned in Sec.~\ref{sec:sec412}, the diversity and dissimilarity of the fakes depend on the distribution in the synthetic dataset for target query. The face resolution and orientation also matter, as small or oblique faces may be missed by the face detector. Finally, users with thousands of friends may consume client storage for face embeddings. It is left as future work to plan a secure client-server protocol for querying friends' face embeddings. Further discussions about our threat model and privacy evaluation are in Supp. C and D.

{\small
\bibliographystyle{ieee_fullname}
\bibliography{egbibnew_}

\begin{thebibliography}{10}\itemsep=-1pt

\bibitem{metadeletes}
Facebook, citing societal concerns, plans to shut down facial recognition
  system.
\newblock
  \url{https://www.nytimes.com/2021/11/02/technology/facebook-facial-recognition.html}.
\newblock Accessed: 2022-03-01.

\bibitem{clearnews}
Facial recognition firm clearview ai says it will soon have 100 billion photos
  in its database to ensure 'almost everyone in the world will be identifiable'
  and wants to expand beyond law enforcement.
\newblock
  \url{https://www.dailymail.co.uk/news/article-10523739/Clearview-AI-seeking-100-billion-photos-facial-recognition-database.html}.
\newblock Accessed: 2022-03-01.

\bibitem{gorilla}
A major flaw in google's algorithm allegedly tagged two black people's faces
  with the word 'gorillas'.
\newblock
  \url{https://www.businessinsider.com/google-tags-black-people-as-gorillas-2015-7}.
\newblock Accessed: 2022-03-01.

\bibitem{clearview}
Clearview ai.
\newblock \url{https://www.clearview.ai/}, 2017.

\bibitem{adjabi2020past}
Insaf Adjabi, Abdeldjalil Ouahabi, Amir Benzaoui, and Abdelmalik Taleb-Ahmed.
\newblock Past, present, and future of face recognition: A review.
\newblock {\em Electronics}, 9(8):1188, 2020.

\bibitem{openface}
Brandon Amos, Bartosz Ludwiczuk, and Mahadev Satyanarayanan.
\newblock Openface: A general-purpose face recognition library with mobile
  applications.
\newblock Technical report, CMU-CS-16-118, CMU School of Computer Science,
  2016.

\bibitem{shai2006}
Shai Avidan and Moshe Butman.
\newblock Efficient methods for privacy preserving face detection.
\newblock {\em Advances in neural information processing systems}, 19, 2006.

\bibitem{gendershades}
Joy Buolamwini and Timnit Gebru.
\newblock Gender shades: Intersectional accuracy disparities in commercial
  gender classification.
\newblock In {\em Conference on fairness, accountability and transparency},
  pages 77--91. PMLR, 2018.

\bibitem{simswap}
Renwang Chen, Xuanhong Chen, Bingbing Ni, and Yanhao Ge.
\newblock Simswap: An efficient framework for high fidelity face swapping.
\newblock In {\em {MM} '20: The 28th {ACM} International Conference on
  Multimedia}, pages 2003--2011. {ACM}, 2020.

\bibitem{chen2021privacy}
Zhenfei Chen, Tianqing Zhu, Ping Xiong, Chenguang Wang, and Wei Ren.
\newblock Privacy preservation for image data: a gan-based method.
\newblock {\em International Journal of Intelligent Systems}, 36(4):1668--1685,
  2021.

\bibitem{ucla}
David Chu, Ilke Demir, Kristen Eichensehr, Jacob~G Foster, Mark~L Green,
  Kristina Lerman, Filippo Menczer, Cailin O’Connor, Edward Parson, Lars
  Ruthotto, et~al.
\newblock White paper: Deep fakery \--- an action plan.
\newblock Technical Report
  \url{http://www.ipam.ucla.edu/wp-content/uploads/2020/01/Whitepaper-Deep-Fakery.pdf},
  Institute for Pure and Applied Mathematics (IPAM), University of California,
  Los Angeles, Los Angeles, CA, Jan. 2020.

\bibitem{8014915}
Anupam Das, Martin Degeling, Xiaoyou Wang, Junjue Wang, Norman Sadeh, and
  Mahadev Satyanarayanan.
\newblock Assisting users in a world full of cameras: A privacy-aware
  infrastructure for computer vision applications.
\newblock In {\em 2017 IEEE Conference on Computer Vision and Pattern
  Recognition Workshops (CVPRW)}, pages 1387--1396, 2017.

\bibitem{mixyn}
Ilke Demir and Umur~A Ciftci.
\newblock Mixsyn: Learning composition and style for multi-source image
  synthesis.
\newblock {\em arXiv preprint arXiv:2111.12705}, 2021.

\bibitem{arcface}
Jiankang Deng, Jia Guo, Xue Niannan, and Stefanos Zafeiriou.
\newblock Arcface: Additive angular margin loss for deep face recognition.
\newblock In {\em CVPR}, 2019.

\bibitem{dong2019efficient}
Yinpeng Dong, Hang Su, Baoyuan Wu, Zhifeng Li, Wei Liu, Tong Zhang, and Jun
  Zhu.
\newblock Efficient decision-based black-box adversarial attacks on face
  recognition.
\newblock In {\em Proceedings of the IEEE/CVF Conference on Computer Vision and
  Pattern Recognition}, pages 7714--7722, 2019.

\bibitem{dumoulin2016learned}
Vincent Dumoulin, Jonathon Shlens, and Manjunath Kudlur.
\newblock A learned representation for artistic style.
\newblock {\em arXiv preprint arXiv:1610.07629}, 2016.

\bibitem{6918642}
Ádám Erdélyi, Tibor Barát, Patrick Valet, Thomas Winkler, and Bernhard
  Rinner.
\newblock Adaptive cartooning for privacy protection in camera networks.
\newblock In {\em 2014 11th IEEE International Conference on Advanced Video and
  Signal Based Surveillance (AVSS)}, pages 44--49, 2014.

\bibitem{gafni2019}
Oran Gafni, Lior Wolf, and Yaniv Taigman.
\newblock Live face de-identification in video.
\newblock In {\em Proceedings of the IEEE/CVF International Conference on
  Computer Vision}, pages 9378--9387, 2019.

\bibitem{timnit2021}
Timnit Gebru, Jamie Morgenstern, Briana Vecchione, Jennifer~Wortman Vaughan,
  Hanna Wallach, Hal~Daum{\'e} Iii, and Kate Crawford.
\newblock Datasheets for datasets.
\newblock {\em Communications of the ACM}, 64(12):86--92, 2021.

\bibitem{goodfellow2014generative}
Ian Goodfellow, Jean Pouget-Abadie, Mehdi Mirza, Bing Xu, David Warde-Farley,
  Sherjil Ozair, Aaron Courville, and Yoshua Bengio.
\newblock Generative adversarial nets.
\newblock {\em Advances in neural information processing systems}, 27, 2014.

\bibitem{gross2006}
Ralph Gross, Latanya Sweeney, Fernando De~la Torre, and Simon Baker.
\newblock Model-based face de-identification.
\newblock In {\em 2006 Conference on computer vision and pattern recognition
  workshop (CVPRW'06)}, pages 161--161. IEEE, 2006.

\bibitem{digitalid}
DGX Digital Identity~Working Group.
\newblock Digital identity in response to covid-19.
\newblock
  \url{https://www.tech.gov.sg/files/media/corporate-publications/FY2021/dgx_2021_digital_identity_in_response_to_covid-19.pdf},
  2021.
\newblock Accessed: 2022-03-01.

\bibitem{insightface}
Jia Guo and Jiankang Deng.
\newblock Insightface.
\newblock \url{https://github.com/deepinsight/insightface}, 2017.

\bibitem{gupta2021}
Aayush Gupta, Ayush Jaiswal, Yue Wu, Vivek Yadav, and Pradeep Natarajan.
\newblock Adversarial mask generation for preserving visual privacy.
\newblock In {\em 2021 16th IEEE International Conference on Automatic Face and
  Gesture Recognition (FG 2021)}, pages 1--5. IEEE, 2021.

\bibitem{huang2017arbitrary}
Xun Huang and Serge Belongie.
\newblock Arbitrary style transfer in real-time with adaptive instance
  normalization.
\newblock In {\em Proceedings of the IEEE international conference on computer
  vision}, pages 1501--1510, 2017.

\bibitem{hukkela2019}
H{\aa}kon Hukkel{\aa}s, Rudolf Mester, and Frank Lindseth.
\newblock Deepprivacy: A generative adversarial network for face anonymization.
\newblock In {\em International symposium on visual computing}, pages 565--578.
  Springer, 2019.

\bibitem{generatedphotos}
Generated~Media Inc.
\newblock Generated photos, 2022.
\newblock Online; accessed 18-February-2022.

\bibitem{Karras_2019_CVPR}
Tero Karras, Samuli Laine, and Timo Aila.
\newblock A style-based generator architecture for generative adversarial
  networks.
\newblock In {\em Proceedings of the IEEE/CVF Conference on Computer Vision and
  Pattern Recognition (CVPR)}, June 2019.

\bibitem{stylegan2}
Tero Karras, Samuli Laine, Miika Aittala, Janne Hellsten, Jaakko Lehtinen, and
  Timo Aila.
\newblock Analyzing and improving the image quality of stylegan.
\newblock In {\em Proceedings of the IEEE/CVF conference on computer vision and
  pattern recognition}, pages 8110--8119, 2020.

\bibitem{dlib}
Davis~E. King.
\newblock Dlib-ml: A machine learning toolkit.
\newblock {\em Journal of Machine Learning Research}, 10:1755--1758, 2009.

\bibitem{kobayashi2014surveillance}
Kento Kobayashi, Keiichi Iwamura, Kitahiro Kaneda, and Isao Echizen.
\newblock Surveillance camera system to achieve privacy protection and crime
  prevention.
\newblock In {\em 2014 Tenth International Conference on Intelligent
  Information Hiding and Multimedia Signal Processing}, pages 463--466. IEEE,
  2014.

\bibitem{korshunov2013using}
Pavel Korshunov and Touradj Ebrahimi.
\newblock Using face morphing to protect privacy.
\newblock In {\em 2013 10th IEEE International Conference on Advanced Video and
  Signal Based Surveillance}, pages 208--213. IEEE, 2013.

\bibitem{korshunova2017}
Iryna Korshunova, Wenzhe Shi, Joni Dambre, and Lucas Theis.
\newblock Fast face-swap using convolutional neural networks.
\newblock In {\em Proceedings of the IEEE international conference on computer
  vision}, pages 3677--3685, 2017.

\bibitem{kortli2020face}
Yassin Kortli, Maher Jridi, Ayman Al~Falou, and Mohamed Atri.
\newblock Face recognition systems: A survey.
\newblock {\em Sensors}, 20(2):342, 2020.

\bibitem{Liao_2018_CVPR}
Fangzhou Liao, Ming Liang, Yinpeng Dong, Tianyu Pang, Xiaolin Hu, and Jun Zhu.
\newblock Defense against adversarial attacks using high-level representation
  guided denoiser.
\newblock In {\em Proceedings of the IEEE Conference on Computer Vision and
  Pattern Recognition (CVPR)}, June 2018.

\bibitem{liu2019few}
Ming-Yu Liu, Xun Huang, Arun Mallya, Tero Karras, Timo Aila, Jaakko Lehtinen,
  and Jan Kautz.
\newblock Few-shot unsupervised image-to-image translation.
\newblock In {\em IEEE International Conference on Computer Vision (ICCV)},
  2019.

\bibitem{liu2017}
Weiyang Liu, Yandong Wen, Zhiding Yu, Ming Li, Bhiksha Raj, and Le Song.
\newblock Sphereface: Deep hypersphere embedding for face recognition.
\newblock In {\em Proceedings of the IEEE conference on computer vision and
  pattern recognition}, pages 212--220, 2017.

\bibitem{ftgan}
Shao-An Lu.
\newblock Ftgan.
\newblock \url{https://github.com/shaoanlu/fewshot-face-translation-GAN}, 2019.

\bibitem{masi2018deep}
Iacopo Masi, Yue Wu, Tal Hassner, and Prem Natarajan.
\newblock Deep face recognition: A survey.
\newblock In {\em 2018 31st SIBGRAPI conference on graphics, patterns and
  images (SIBGRAPI)}, pages 471--478. IEEE, 2018.

\bibitem{maximov2020}
Maxim Maximov, Ismail Elezi, and Laura Leal-Taix{\'e}.
\newblock Ciagan: Conditional identity anonymization generative adversarial
  networks.
\newblock In {\em Proceedings of the IEEE/CVF Conference on Computer Vision and
  Pattern Recognition}, pages 5447--5456, 2020.

\bibitem{blaz2021}
Bla{\v{z}} Meden, Peter Rot, Philipp Terh{\"o}rst, Naser Damer, Arjan Kuijper,
  Walter~J Scheirer, Arun Ross, Peter Peer, and Vitomir {\v{S}}truc.
\newblock Privacy--enhancing face biometrics: A comprehensive survey.
\newblock {\em IEEE Transactions on Information Forensics and Security}, 2021.

\bibitem{microsoftfacerec}
Microsoft.
\newblock Face api.
\newblock
  \url{https://azure.microsoft.com/en-us/services/cognitive-services/face/#overview/},
  2017.

\bibitem{newton2005}
Elaine~M Newton, Latanya Sweeney, and Bradley Malin.
\newblock Preserving privacy by de-identifying face images.
\newblock {\em IEEE transactions on Knowledge and Data Engineering},
  17(2):232--243, 2005.

\bibitem{nightingale2021synthetic}
Sophie Nightingale, Shruti Agarwal, Erik H{\"a}rk{\"o}nen, Jaakko Lehtinen, and
  Hany Farid.
\newblock Synthetic faces: how perceptually convincing are they?
\newblock {\em Journal of Vision}, 21(9):2015--2015, 2021.

\bibitem{fsgan}
Yuval Nirkin, Yosi Keller, and Tal Hassner.
\newblock {FSGAN}: Subject agnostic face swapping and reenactment.
\newblock In {\em Proceedings of the IEEE International Conference on Computer
  Vision}, pages 7184--7193, 2019.

\bibitem{faceswap}
Yuval Nirkin, Iacopo Masi, Anh~Tuan Tran, Tal Hassner, and G\'{e}rard Medioni.
\newblock On face segmentation, face swapping, and face perception.
\newblock In {\em IEEE Conference on Automatic Face and Gesture Recognition},
  2018.

\bibitem{oh2016}
Seong~Joon Oh, Rodrigo Benenson, Mario Fritz, and Bernt Schiele.
\newblock Faceless person recognition: Privacy implications in social media.
\newblock In {\em European Conference on Computer Vision}, pages 19--35.
  Springer, 2016.

\bibitem{oh2017}
Seong~Joon Oh, Mario Fritz, and Bernt Schiele.
\newblock Adversarial image perturbation for privacy protection a game theory
  perspective.
\newblock In {\em 2017 IEEE International Conference on Computer Vision
  (ICCV)}, pages 1491--1500. IEEE, 2017.

\bibitem{park2019SPADE}
Taesung Park, Ming-Yu Liu, Ting-Chun Wang, and Jun-Yan Zhu.
\newblock Semantic image synthesis with spatially-adaptive normalization.
\newblock In {\em Proceedings of the IEEE Conference on Computer Vision and
  Pattern Recognition}, 2019.

\bibitem{proencca2021uu}
Hugo Proen{\c{c}}a.
\newblock The uu-net: Reversible face de-identification for visual surveillance
  video footage.
\newblock {\em IEEE Transactions on Circuits and Systems for Video Technology},
  32(2):496--509, 2021.

\bibitem{ren2018}
Zhongzheng Ren, Yong~Jae Lee, and Michael~S Ryoo.
\newblock Learning to anonymize faces for privacy preserving action detection.
\newblock In {\em Proceedings of the european conference on computer vision
  (ECCV)}, pages 620--636, 2018.

\bibitem{8014911}
Natacha Ruchaud and Jean-Luc Dugelay.
\newblock Aseppi: Robust privacy protection against de-anonymization attacks.
\newblock In {\em 2017 IEEE Conference on Computer Vision and Pattern
  Recognition Workshops (CVPRW)}, pages 1352--1359, 2017.

\bibitem{facenet}
Florian Schroff, Dmitry Kalenichenko, and James Philbin.
\newblock Facenet: A unified embedding for face recognition and clustering.
\newblock In {\em Proceedings of the IEEE conference on computer vision and
  pattern recognition}, pages 815--823, 2015.

\bibitem{shan2020}
Shawn Shan, Emily Wenger, Jiayun Zhang, Huiying Li, Haitao Zheng, and Ben~Y
  Zhao.
\newblock Fawkes: Protecting privacy against unauthorized deep learning models.
\newblock In {\em 29th USENIX Security Symposium (USENIX Security 20)}, pages
  1589--1604, 2020.

\bibitem{shen2020interpreting}
Yujun Shen, Jinjin Gu, Xiaoou Tang, and Bolei Zhou.
\newblock Interpreting the latent space of gans for semantic face editing.
\newblock In {\em Proceedings of the IEEE/CVF Conference on Computer Vision and
  Pattern Recognition}, pages 9243--9252, 2020.

\bibitem{shirai2019}
Sola Shirai and Jacob Whitehill.
\newblock Privacy-preserving annotation of face images through
  attribute-preserving face synthesis.
\newblock In {\em Proceedings of the IEEE/CVF Conference on Computer Vision and
  Pattern Recognition Workshops}, pages 0--0, 2019.

\bibitem{sun2018cvpr}
Qianru Sun, Liqian Ma, Seong~Joon Oh, Luc Van~Gool, Bernt Schiele, and Mario
  Fritz.
\newblock Natural and effective obfuscation by head inpainting.
\newblock In {\em Proceedings of the IEEE Conference on Computer Vision and
  Pattern Recognition}, pages 5050--5059, 2018.

\bibitem{sun2018eccv}
Qianru Sun, Ayush Tewari, Weipeng Xu, Mario Fritz, Christian Theobalt, and
  Bernt Schiele.
\newblock A hybrid model for identity obfuscation by face replacement.
\newblock In {\em Proceedings of the European Conference on Computer Vision
  (ECCV)}, pages 553--569, 2018.

\bibitem{deepid}
Yi Sun, Xiaogang Wang, and Xiaoou Tang.
\newblock Deep learning face representation from predicting 10,000 classes.
\newblock In {\em Proceedings of the IEEE Conference on Computer Vision and
  Pattern Recognition (CVPR)}, June 2014.

\bibitem{deepface}
Yaniv Taigman, Ming Yang, Marc'Aurelio Ranzato, and Lior Wolf.
\newblock Deepface: Closing the gap to human-level performance in face
  verification.
\newblock In {\em Proceedings of the IEEE conference on computer vision and
  pattern recognition}, pages 1701--1708, 2014.

\bibitem{tanisik2016facial}
Gokhan Tanisik, Cemil Zalluhoglu, and Nazli Ikizler-Cinbis.
\newblock Facial descriptors for human interaction recognition in still images.
\newblock {\em Pattern Recognition Letters}, 73:44--51, 2016.

\bibitem{thies2019deferred}
Justus Thies, Michael Zollh{\"o}fer, and Matthias Nie{\ss}ner.
\newblock Deferred neural rendering: Image synthesis using neural textures.
\newblock {\em ACM Transactions on Graphics (TOG)}, 38(4):1--12, 2019.

\bibitem{viola2001rapid}
Paul Viola and Michael Jones.
\newblock Rapid object detection using a boosted cascade of simple features.
\newblock In {\em Proceedings of the 2001 IEEE computer society conference on
  computer vision and pattern recognition. CVPR 2001}, volume~1, pages I--I.
  Ieee, 2001.

\bibitem{vishwamitra2017}
Nishant Vishwamitra, Bart Knijnenburg, Hongxin Hu, Yifang~P Kelly~Caine, et~al.
\newblock Blur vs. block: Investigating the effectiveness of privacy-enhancing
  obfuscation for images.
\newblock In {\em Proceedings of the IEEE Conference on Computer Vision and
  Pattern Recognition Workshops}, pages 39--47, 2017.

\bibitem{wang2018additive}
Feng Wang, Jian Cheng, Weiyang Liu, and Haijun Liu.
\newblock Additive margin softmax for face verification.
\newblock {\em IEEE Signal Processing Letters}, 25(7):926--930, 2018.

\bibitem{wen2021identitydp}
Yunqian Wen, Li Song, Bo Liu, Ming Ding, and Rong Xie.
\newblock Identitydp: Differential private identification protection for face
  images.
\newblock {\em arXiv preprint arXiv:2103.01745}, 2021.

\bibitem{we2016m}
Yandong Wen, Kaipeng Zhang, Zhifeng Li, and Yu Qiao.
\newblock A discriminative feature learning approach for deep face recognition.
\newblock In {\em European conference on computer vision}, pages 499--515.
  Springer, 2016.

\bibitem{wu2019}
Yifan Wu, Fan Yang, Yong Xu, and Haibin Ling.
\newblock Privacy-protective-gan for privacy preserving face de-identification.
\newblock {\em Journal of Computer Science and Technology}, 34(1):47--60, 2019.

\bibitem{yuan2017image}
Lin Yuan and Touradj Ebrahimi.
\newblock Image privacy protection with secure jpeg transmorphing.
\newblock {\em IET Signal Processing}, 11(9):1031--1038, 2017.

\bibitem{NEURIPS2019_d8700cbd}
Haichao Zhang and Jianyu Wang.
\newblock Defense against adversarial attacks using feature scattering-based
  adversarial training.
\newblock In H. Wallach, H. Larochelle, A. Beygelzimer, F. d\textquotesingle
  Alch\'{e}-Buc, E. Fox, and R. Garnett, editors, {\em Advances in Neural
  Information Processing Systems}, volume~32. Curran Associates, Inc., 2019.

\bibitem{zhao2003face}
Wenyi Zhao, Rama Chellappa, P~Jonathon Phillips, and Azriel Rosenfeld.
\newblock Face recognition: A literature survey.
\newblock {\em ACM computing surveys (CSUR)}, 35(4):399--458, 2003.

\bibitem{zhong2020towards}
Yaoyao Zhong and Weihong Deng.
\newblock Towards transferable adversarial attack against deep face
  recognition.
\newblock {\em IEEE Transactions on Information Forensics and Security},
  16:1452--1466, 2020.

\bibitem{zhu2019}
Chen Zhu, W~Ronny Huang, Hengduo Li, Gavin Taylor, Christoph Studer, and Tom
  Goldstein.
\newblock Transferable clean-label poisoning attacks on deep neural nets.
\newblock In {\em International Conference on Machine Learning}, pages
  7614--7623. PMLR, 2019.

\bibitem{zou2018}
James Zou and Londa Schiebinger.
\newblock Ai can be sexist and racist—it’s time to make it fair, 2018.

\end{thebibliography}
}

\title{My Face My Choice: Privacy Enhancing Deepfakes for \\ Social Media Anonymization
\\Supplementary Material}

\author{Umur A. \c{C}ift\c{c}i\\
Binghamton University \\
{\tt\small uciftci@binghamton.edu}
\and
Gokturk Yuksek \\
Binghamton University \\
{\tt\small gokturk@binghamton.com}
\and
\.Ilke Demir \\
Intel Labs \\
{\tt\small ilke.demir@intel.com}
}

\maketitle
\appendix
\section{Comparison}
In Fig.~\ref{fig:additional}, we perform experiments of CIAGAN~\cite{maximov2020} and DeepPrivacy~\cite{hukkela2019} with MFMC using their sample images. We observe that faces produced by MFMC are much coherent in skin and gender attributes, preserve the expression better, and overall provide more realistic results. Remark that, MFMC has the goal of creating quantitatively dissimilar and realistic deepfakes, so it has more relaxed constraints on preserving the identity of the target and more strict constraints on the image and expression quality. 
%Moreover, those approaches are much more time consuming (X minutes and Y minutes to create one deepfake), whereas MFMC transforms a photo with ten faces under a minute.
\begin{figure*}[ht]
\centering
\includegraphics[width=1\linewidth]{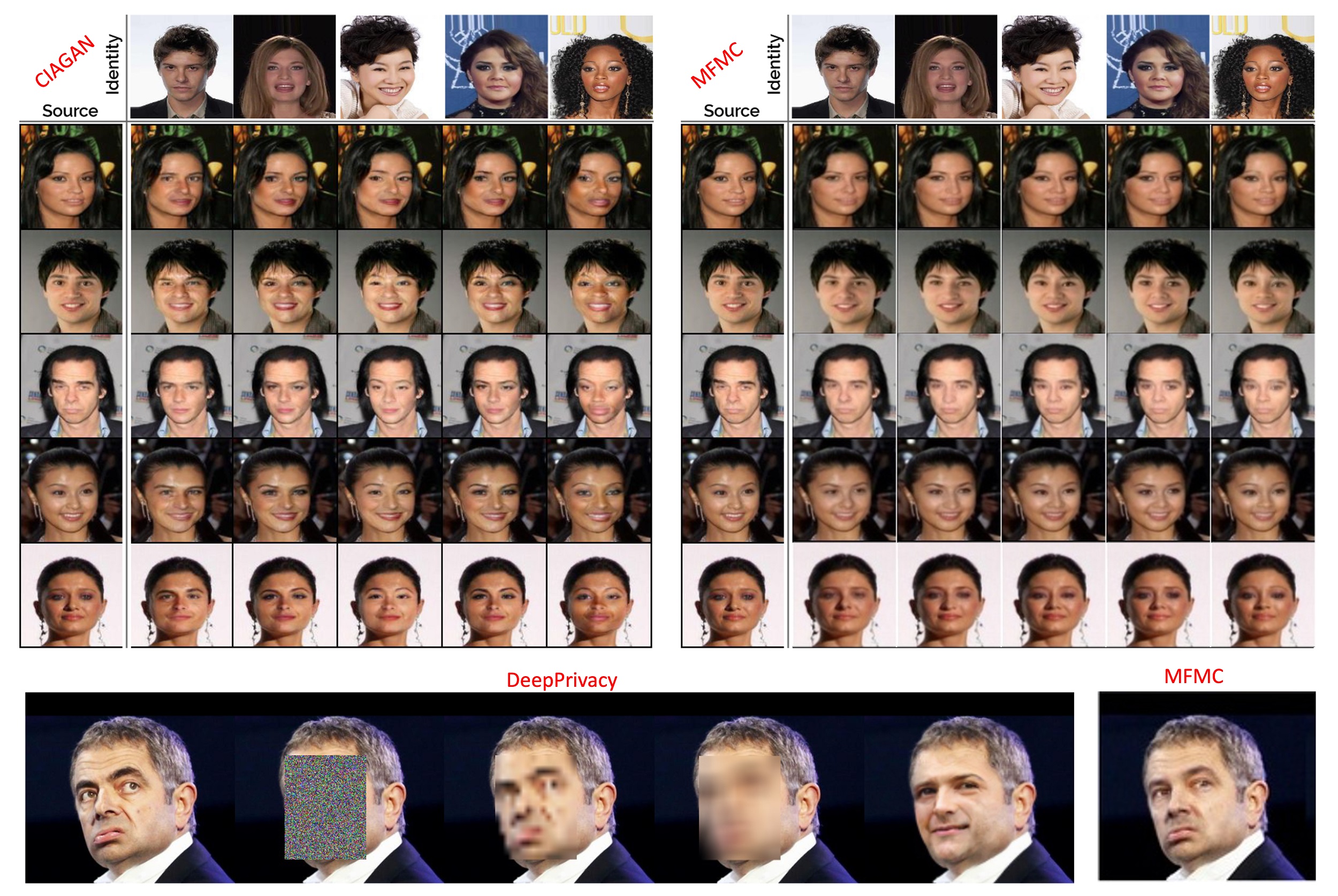} % 
\caption{\textbf{Comparison.} We replicate the results of CIAGAN~\cite{maximov2020} and DeepPrivacy~\cite{hukkela2019} (left) using MFMC. Artifacts from skin color, gender difference, and other subtle differences are not observed in MFMC results.}
\label{fig:additional}
\end{figure*}

\section{Additional Face Recognition Accuracies}
We extend the exploration of how much MFMC can trick face recognition approaches if we use SSIM and RMSE similarity metrics for target face query in Tab.~\ref{tab:facerecSSIM}. Similar to the results using the face embedding similarity, MFMC can trick 71\% on average for SSIM, and 77\% for RMSE. Although incorrect recognition rate is higher, we use the embedding distance as grounded in the main paper. Moreover, the embedding space resembles the latent space learned by the face recognition model more than RMSE and SSIM spaces, thus better ``tricking'' is not surprising.
\begin{table}[ht]
\begin{tabular}{c|cc|cc}\centering
%\multirow{Face} & \multicolumn{2}{c|}{Source - Target} & \multicolumn{2}{c}{Source - Result} \\  \cline{2-5} Detector 
Face & \multicolumn{2}{c|}{Source vs. Target} & \multicolumn{2}{c}{Source vs. Result} \\  \cline{2-5} Detector 
          & \multicolumn{1}{c|}{SSIM}     & RMSE        & \multicolumn{1}{c|}{SSIM}    & RMSE    \\ \hline
FaceNet512& \multicolumn{1}{c|}{0.001}        & 0.0     & \multicolumn{1}{c|}{0.12}        & 0.16          \\
OpenFace  & \multicolumn{1}{c|}{0.001}        & 0.003   & \multicolumn{1}{c|}{0.17}        & 0.24           \\
FaceNet   & \multicolumn{1}{c|}{0.03}        & 0.02     & \multicolumn{1}{c|}{0.27}        & 0.34           \\
DLib      & \multicolumn{1}{c|}{0.02}        & 0.05     & \multicolumn{1}{c|}{0.30}         & 0.44          \\
ArcFace   & \multicolumn{1}{c|}{0.05}        & 0.03     & \multicolumn{1}{c|}{0.29}         & 0.45           \\ 
DeepID    & \multicolumn{1}{c|}{0.005}        & 0.01    & \multicolumn{1}{c|}{0.44}         & 0.52          \\
DeepFace  & \multicolumn{1}{c|}{0.03}        & 0.06     & \multicolumn{1}{c|}{0.47}         & 0.53          \\\hline
Average   & \multicolumn{1}{c|}{0.02}        & 0.01     & \multicolumn{1}{c|}{0.29}         & 0.23         
\end{tabular}
\caption{Seven SOTA face recognition approaches are compared on MFMC results based on face identification accuracy, where the furthest face is chosen in SSIM and RMSE metric spaces.}\label{tab:facerecSSIM}
\end{table}
\\ 

Switching from cosine to $L_2$ distance for embedding comparisons, Tab.~\ref{tab:facerec} documents face recognition results where MFMC is able to reduce the accuracy to 49\% on the average (last), and to 44\% if we lift the randomness (third).
\begin{table}[ht]
\begin{tabular}{c|cc|cc}\centering
%\multirow{Face} & \multicolumn{2}{c|}{Source - Target} & \multicolumn{2}{c}{Source - Result} \\  \cline{2-5} Detector 
Face & \multicolumn{2}{c|}{Source vs. Target} & \multicolumn{2}{c}{Source vs. Result} \\  \cline{2-5} Detector 
                               & \multicolumn{1}{c|}{Furthest}     & Random    & \multicolumn{1}{c|}{Furthest}    & Random    \\ \hline
FaceNet512                     & \multicolumn{1}{c|}{0.08}        & 0.05      & \multicolumn{1}{c|}{0.34}        & 0.43      \\
OpenFace                       & \multicolumn{1}{c|}{0.005}       & 0.009     & \multicolumn{1}{c|}{0.31}        & 0.34      \\
FaceNet                        & \multicolumn{1}{c|}{0.08}        & 0.06      & \multicolumn{1}{c|}{0.40}        & 0.43      \\
DLib                           & \multicolumn{1}{c|}{0.05}        & 0.10      & \multicolumn{1}{c|}{0.50}        & 0.64      \\
ArcFace                        & \multicolumn{1}{c|}{0.02}        & 0.03      & \multicolumn{1}{c|}{0.28}        & 0.35       \\ 
DeepID                         & \multicolumn{1}{c|}{0.009}       & 0.01      & \multicolumn{1}{c|}{0.58}        & 0.58      \\
DeepFace                       & \multicolumn{1}{c|}{0.22}        & 0.27      & \multicolumn{1}{c|}{0.65}        & 0.66      \\\hline
Average                        & \multicolumn{1}{c|}{0.07}        & 0.07      & \multicolumn{1}{c|}{0.44}        & 0.49     
\end{tabular}
\caption{\textbf{Face Recognition Accuracies after MFMC.} Seven SOTA face recognition approaches are compared on MFMC results, using $L_2$ distance between face embeddings.}\label{tab:facerec}
\end{table}

\section{Threat Model}
Our main threat model is automatic face recognition systems that associate faces with personally identifiable information or assign permanent face embeddings. CCTV cameras, millions of images on social media, and constantly evolving media sources capture all of us in some photos voluntarily or involuntarily. We would like to disable attackers to mine identity information from these photos, while enabling willing users to participate in the social platform.

For users who grant no access or for non-users, their identity never gets associated with the photo, no embedding is generated and the real face is disposed from the client right after being deepfaked at upload time -- assuming there is no interruption at upload time. For users with other options, their face embeddings are shared only with friends in an encrypted way, and instances of their real faces are stored on the server. This requires trusting the social media platform for handling the process privately without human intervention, for having a secure client-server transmission protocol, and for not leaking the photos or embeddings.

% We understand that just changing the face is not fully safe as long as the original photo can still be linked to personal information. We also believe that completely removing these faces or not protecting certain expressions would make the picture not look real, and as a result the platform will be less attractive. To the best of our knowledge MFMC is the first system to take privacy enhancing anonyimization to the next level by asking how such a system can be implemented, what options should be available and in what manner it should be used. Compared to not using any, we believe our method will add more cumber to the human attacker in order to recognize the identity of a person, and give good protection against web crawlers who search for the person's identification through faces.
% In terms of age and gender attribute preservation, they are completely optional and the user does not depend on it, but by creating a similar face creates more realistic and believe image compared to the non similar faces and in our view that what most of the social media posters care(having a good looking image).

\section{Privacy Evaluation}
In contrast to anonymization methods which aggregates data points into groups that disable inferring individual information, our approach masks each face with a deepfake per photo. These deepfakes should not even be considered as quasi-identifiers, as they no longer preserve the identity. Having access to $k-1$ deepfake versions of the same face does not enable reconstructing the original face, even if the original photo is in that set (without being known as the original), which satisfies \textit{k-anonymity}. The age and gender groups to create deepfakes are synthetic calculations that we do not seek the exact values for, they are approximate ranges to preserve the photorealism. 

On the other hand we are vulnerable to linkability attacks if the same image is posted on a platform without anonymization, or if there is personally identifiable data in the image in another form than faces.

\end{document}